\documentclass[final,12pt]{elsarticle}

\usepackage{graphicx}
\usepackage{xcolor}
\usepackage{multirow}

\usepackage{hyperref}
\usepackage{microtype}

\usepackage{color}
\usepackage{soul}
\usepackage{subcaption}
\usepackage{amssymb}
\usepackage{makeidx}  
\usepackage{caption}
\usepackage{algorithm, algpseudocode}
\usepackage{amsmath}
\usepackage{cleveref}

\usepackage{tikz}
\usepackage{tikz-3dplot}
\usepackage{pgfplots}
\usepackage{pgfplotstable, filecontents}

\DeclareMathOperator*{\argmax}{arg\,max}

\usepackage{hyperref}
\hypersetup{
    colorlinks = true,
    citecolor = {blue}
}
\usepackage{booktabs}       
\usepackage{amsfonts}       
\usepackage{bm}             

\biboptions{authoryear}

\newcommand\blfootnote[1]{%
  \begingroup
  \renewcommand\thefootnote{}\footnote{#1}%
  \addtocounter{footnote}{-1}%
  \endgroup
}

\journal{Expert Systems with Applications}

\hypersetup{draft}

\begin{document}

\begin{frontmatter}

\title{Tensor Analysis with n-Mode Generalized Difference Subspace}



\author{Bernardo Bentes Gatto\corref{cor1}}
\address{Center for Artificial Intelligence Research (C-AIR)\\ Tsukuba, Japan \\ \url{bernardo@cvlab.cs.tsukuba.ac.jp}}
\address{Federal University of Amazonas\\ Institute of Computing \\
Manaus, AM, Brazil}

\author{Eulanda Miranda dos Santos}                 
\address{Federal University of Amazonas\\ Institute of Computing \\
Manaus, AM, Brazil\\
\url{emsantos@icomp.ufam.edu.br}}

\author{Alessandro Lameiras Koerich}                     
\address{\'{E}cole de Technologie Sup\'{e}rieure, Universit\'{e} du Qu\'{e}bec \\ 
Department of Software and IT Engineering\\ Montreal, QC, Canada \\ \url{alessandro.koerich@etsmtl.ca}}

\author{Kazuhiro Fukui}                            
\address{Center for Artificial Intelligence Research (C-AIR)\\ Tsukuba, Japan \\ \url{kfukui@cs.tsukuba.ac.jp}}

\author{Waldir Sabino da Silva J\'{u}nior}
\address{Federal University of Amazonas\\ Institute of Computing \\
Manaus, AM, Brazil \\ \url{waldirjr@ufam.edu.br}}

\cortext[cor1]{Corresponding author}
\newpageafter{author}

\begin{abstract}

The increasing use of multiple sensors, which produce a large amount of multi-dimensional data, requires efficient representation and classification methods. In this paper, we present a new method for multi-dimensional data classification that relies on two premises: 1) multi-dimensional data are usually represented by tensors, since this brings benefits from multilinear algebra and established tensor factorization methods; and 2) multilinear data can be described by a subspace of a vector space. The subspace representation has been employed for pattern-set recognition, and its tensor representation counterpart is also available in the literature. However, traditional methods do not use discriminative information of the tensors, degrading the classification accuracy. In this case, generalized difference subspace (GDS) provides an enhanced subspace representation by reducing data redundancy and revealing discriminative structures. Since GDS does not handle tensor data, we propose a new projection called n-mode GDS, which efficiently handles tensor data. We also introduce the n-mode Fisher score as a class separability index and an improved metric based on the geodesic distance for tensor data similarity. The experimental results on gesture and action recognition show that the proposed method outperforms methods commonly used in the literature without relying on pre-trained models or transfer learning.
\end{abstract}

\begin{keyword}
	Action recognition; Tensor data classification; Generalized difference subspace; n-mode singular value decomposition.
\end{keyword}

\blfootnote{Abbreviations: 
DCC: discriminative canonical correlation; 
GDS: generalized difference subspace; 
TCCA: tensor canonical correlation analysis; 
MPCA: multilinear  principal  component  analysis; 
MLDA: multilinear linear discriminant analysis; 
PGM: product Grassmann manifold; 
TB: tangent bundle; 
DNN: deep neural network; 
MSM: mutual subspace method; 
CNN: convolutional neural network; 
SVD: single value decomposition; 
DS: difference subspace; 
MDS: multi dimensional scaling;
LRNTF: laplacian nonnegative tensor factorization with regularization.}

\end{frontmatter}



\section{Introduction}
\label{S:Introduction}

Many applications make use of multi-dimensional data, such as multiple-view image recognition and video analysis. Due to the increasing data density produced by sensors, improved techniques are needed in order to process this kind of data. Tensors, which can be defined as a generalization of matrices, are a suitable model for such data representation since tensors allow a natural representation of multi-dimensional data. For instance, video data are intuitively described by their correlated images over the time axis. Through vectorization and concatenation of the video data pixels, it is possible to produce a representation that describes the data as a matrix or a vector. This vectorized representation can then be used to build a machine learning model. However, this representation does not provide an intuitive or natural pattern representation. Furthermore, the vectorization procedure may degrade the spatio-temporal relationship between pixels of video tensor data, causing information loss~\citep{VEC1,VEC2}.

Applications that benefit from tensorial representation include high-resolution video analysis, hyperspectral image classification, medical image analysis, gene expression representation and recommendation systems~\citep{GEN1,GEN2,HIG1,HIG2,HYP1,HYP2,MED1,MED2,REC1_exp,REC2_exp}. For example, bioelectrical time signals~\citep{EEG1_exp,EEG2_exp} are usually obtained by sensors based on differential amplifiers, which record the signal difference between two electrodes connected to the skin where the signal difference changes over time. Since many sensors can be used to cover a wider area, this data acquisition produces a massive number of time-varying signals, where not only the temporal correlation but also the spatial structure between the collected signals need to be used. Another recent example is self-driving cars that are equipped with multiple sensors and produce a large amount of data, which requires an efficient representation~\citep{CAR1,CAR2}. Finally, the use of non-efficient methods to handle high-dimensional data can reduce the associated hardware cost.

Currently, training deep neural network (DNN) architectures from scratch is not feasible for handling $3$-mode or higher mode tensors when datasets have a small number of samples. More precisely, training a DNN requires large datasets due to the large number of trainable parameters. Furthermore, the computational complexity for training a DNN architecture may increase exponentially with the number of modes, which requires an increasing amount of data and computational resources, restricting the range of possible applications. To overcome this problem, we propose a method in which the complexity grows linearly with the number of tensor modes, making the proposed method an alternative for tensor data classification.

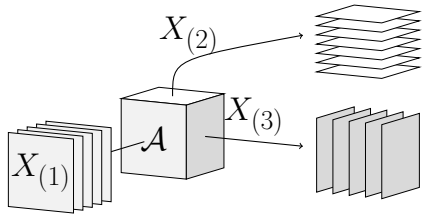
\begin{figure}[t!]
	\centering
	\scalebox{.5}{%
	\tdplotsetmaincoords{80}{120}

\begin{tikzpicture}[tdplot_main_coords]

\coordinate (O) at (0,0,0);

\draw[fill=white]    (0,0,2) -- (0,2,2) -- (2,2,2) -- (2,0,2) -- cycle;
\draw[fill=gray!10]  (2,2,2) -- (2,2,0) -- (2,0,0) -- (2,0,2) -- cycle;
\draw[fill=gray!30]  (0,2,2) -- (2,2,2) -- (2,2,0) -- (0,2,0) -- cycle;


\draw[thick, ->] (1,2,1) -- node[midway,above] {\huge $X_{(3)}$} (1,5,1);
\draw[thick, ->] (2.5,1,1) -- node[midway,above] {} (5,1,1);
\draw[thick, ->] (1,1,2) .. controls (1,1,3) .. node[midway,above] {\huge $X_{(2)}$} (1,5,4);

\draw[fill=gray!10, opacity=1]  (6,2,2) -- (6,2,0) -- (6,0,0) -- (6,0,2) -- cycle;
\draw[fill=gray!10, opacity=1]  (6.5,2,2) -- (6.5,2,0) -- (6.5,0,0) -- (6.5,0,2) -- cycle;
\draw[fill=gray!10, opacity=1]  (7,2,2) -- (7,2,0) -- (7,0,0) -- (7,0,2) -- cycle;
\draw[fill=gray!10, opacity=1]  (7.5,2,2) -- (7.5,2,0) -- (7.5,0,0) -- (7.5,0,2) -- cycle;
\draw[fill=gray!10, opacity=1]  (8,2,2) -- (8,2,0) -- (8,0,0) -- (8,0,2) -- cycle;

\draw[fill=gray!30, opacity=1]  (2,6,2) -- (2,6,0) -- (0,6,0) -- (0,6,2) -- cycle;
\draw[fill=gray!30, opacity=1]  (2,6.5,2) -- (2,6.5,0) -- (0,6.5,0) -- (0,6.5,2) -- cycle;
\draw[fill=gray!30, opacity=1]  (2,7,2) -- (2,7,0) -- (0,7,0) -- (0,7,2) -- cycle;
\draw[fill=gray!30, opacity=1]  (2,7.5,2) -- (2,7.5,0) -- (0,7.5,0) -- (0,7.5,2) -- cycle;
\draw[fill=gray!30, opacity=1]  (2,8,2) -- (2,8,0) -- (0,8,0) -- (0,8,2) -- cycle;

\draw[fill=white, opacity=1]  (0,6,3.25) -- (0,8,3.25) -- (2,8,3.25) -- (2,6,3.25) -- cycle;
\draw[fill=white, opacity=1]  (0,6,3.5) -- (0,8,3.5) -- (2,8,3.5) -- (2,6,3.5) -- cycle;
\draw[fill=white, opacity=1]  (0,6,3.75) -- (0,8,3.75) -- (2,8,3.75) -- (2,6,3.75) -- cycle;
\draw[fill=white, opacity=1]  (0,6,4) -- (0,8,4) -- (2,8,4) -- (2,6,4) -- cycle;
\draw[fill=white, opacity=1]  (0,6,4.25) -- (0,8,4.25) -- (2,8,4.25) -- (2,6,4.25) -- cycle;
\draw[fill=white, opacity=1]  (0,6,4.5) -- (0,8,4.5) -- (2,8,4.5) -- (2,6,4.5) -- cycle;
\draw[fill=white, opacity=1]  (0,6,4.75) -- (0,8,4.75) -- (2,8,4.75) -- (2,6,4.75) -- cycle;

\node (A) at (8,1,1) {\huge $X_{(1)}$};

\node (A) at (2,1,1) {\Huge $\mathcal{A}$};

\end{tikzpicture}}%
	\caption[Illustration of the unfolding procedure of a $3$-mode tensor.]{Illustration of the unfolding procedure of a $3$-mode tensor. The unfolding of the $3$-mode tensor $\mathcal A$ produces $3$ sets of matrices $X_{(1)}$, $X_{(2)}$ and $X_{(3)}$.}
	\label{fig:unfolding}
\end{figure}

The order of a tensor is related to the number of dimensions, also known as ways or modes~\citep{MODE}. Tensor unfolding is a procedure that reorganizes tensor data in such a way that each mode can be analyzed separately, possibly revealing correlations that were not immediately observed. For example, video data may provide a $3$-mode tensor, consisting of $2$ spatial modes and a temporal one. In terms of tensor unfolding, this kind of $3$-mode tensor can be represented by $3$ subspaces, where each subspace is computed from one of the tensor-unfolded modes. This tensor-unfolding procedure is shown in Figure~\ref{fig:unfolding}. The tensor-unfolding technique is relevant for the interpretability of the modes, as it is in the case of medical image analysis~\citep{Interp_Med1,Interp_Med2}.

In computer vision, subspaces are systematically employed to express pattern-set data. A pattern-set can be defined as a collection of observations that share a particular label. The mutual subspace method (MSM) was introduced to solve pattern-set recognition problems~\citep{FUKUI_1_subspace,MAEDA_1_subspace}. The central hypothesis of MSM is that a pattern-set produces a compact cluster in a high-dimensional vector space, which can be efficiently represented by a subspace computed by eigen-decomposition, for instance. In MSM, the similarity between subspaces is computed from the canonical angles~\citep{CAN1,CAN2}. The advantages of using MSM include its extremely compact representation and its robustness to noise. For example, it is reasonable that $20\%$ of the basis vectors generated by eigen-decomposition can efficiently represent $90\%$ of a particular pattern-set. The product Grassmann manifold (PGM) is one example of the use of subspaces to represent tensor data such as in action recognition problems~\citep{LUI1,LUI2}. In this case, PGM extracts subspaces from tensor data and represents them as a point on the product space of $3$ Grassmann manifolds, where each subspace corresponds to a point on one of the Grassmann manifolds. It is worth noting that PGM is capable of handling more than $3$ modes, although the application of action recognition requires only $3$ modes. The classification is then performed based on the chordal distance~\citep{Chordal_distance1,Chordal_distance2} on the product manifold.

It is known that the chordal distance on a product manifold is equivalent to the Cartesian product of geodesics from the manifolds~\citep{ARG1,ARG2}. By incorporating the chordal distance on the product manifold, we may express the relation between the subspaces of all available tensor modes in a unified design. The nearest neighbor classifier using the chordal distance on a Grassmann manifold is equivalent to MSM. According to this theoretical relation, PGM and MSM are equivalent in terms of pattern-set representation. Although MSM has been established as a standard framework in the research field of pattern-set recognition, solving many practical problems, its discriminant ability is known to be insufficient since each class subspace is created without reflecting the between-class relations. Therefore, PGM inherits the main disadvantage of MSM, namely, the absence of a discriminative mechanism.

\citet{CMSM} proposed the concept of the difference subspace (DS), which minimizes data redundancy while extracting suitable features for classification. Further refinement of this method was introduced to handle multiple class subspaces by using the generalized difference subspace (GDS). GDS was then employed to solve several pattern-set related problems, such as face and handshape classification. More precisely, GDS projection acts as a feature extractor for MSM. Since GDS represents the difference among class subspaces, the GDS projection can increase the angles among the class subspaces closer to orthogonal. As a result, GDS projection operates as a quasi-orthogonalization process, which is a practical feature extraction for any subspace-based method. These operations allow the generation of discriminative features, overcoming the limitations of MSM. Despite its useful properties, GDS has not been yet employed in tensor data applications since, in such applications, the ordering relations between patterns must be preserved. Since GDS is based on the eigen-decomposition of the pattern-sets, temporal relations between the patterns are usually lost.

It is worth mentioning that tensor learning paradigms other than supervised learning are discussed in the literature. One example of unsupervised learning is Laplacian nonnegative tensor factorization with regularization (LRNTF), introduced by~\citet{LRNTF} for image representation. This method considers the feature space underlying manifold structure and preserves the geometric relation between data points. The proposed tensor factorization method is then evaluated in a clustering task. LRNTF is applied in a k-means clustering algorithm, which gives advantages when the dimensionality of the feature space is high. The interested reader can refer to~\cite{tensor_review} for a comprehensive list of recent advances in tensor analysis.

It is worth noting that eigen-decomposition does not preserve the ordering of feature vectors in a set. For a clear example, let us consider two ordered feature sets $X_{(1)}=\{x_1,~x_2,~x_3,\ldots,x_k\}$ and $X_{(1)}^{\,\prime}=\{x_k,~x_{k-1},~x_{k-2},\ldots,x_1\}$. Their non-centered autocorrelation matrices $A_{{(1)}}$ and $A_{(1)}^{\,\prime}$ represent the same entries since $A_{{(1)}} = x_1^{\top}x_1 + x_2^{\top}x_2 + x_3^{\top}x_3 +\ldots + x_k^{\top}x_k$ and $A_{(1)}^{\,\prime} = x_k^{\top}x_k + x_{(k-1)}^{\top}x_{(k-1)} + x_{(k-2)}^{\top}x_{(k-2)} +\ldots + x_1^{\top}x_1$, and therefore $A_{{(1)}} = A_{(1)}^{\,\prime}$ due to the commutative property of addition. Since $A_{{(1)}} = A_{(1)}^{\,\prime}$, their eigen-decompositions  produce the same basis vectors.

Although employing the unfolding procedure described in Figure~\ref{fig:unfolding} does not change the basis vectors of $A_{{(1)}}$ and $A_{(1)}^{\,\prime}$, the unfolded features of $X_{{(2)}}$ and $X_{(2)}^{\,\prime}$ represent their feature entries modified exactly in the ordering of $X_{{(1)}}$ and $X_{(1)}^{\,\prime}$, which produces distinct non-centered autocorrelation matrices, thus providing dissimilar basis vectors and preserving the ordering of the pattern-set.

In this paper, we introduce the $n$-mode GDS projection, which is able to extract discriminative information from tensor data and to provide suitable subspaces for tensor data classification. Under the n-mode formulation, we can efficiently express tensor data as a point on a product manifold~\citep{LUI1,LUI2}, simplifying the tensorial data representation as well as inheriting the main characteristics of GDS. In order to evaluate the quasi-orthogonality properties of the proposed method, we develop a new separability index based on the Fisher score. Since the Fisher score is not able to handle tensorial data, we redefine its traditional formulation to handle tensor data. Once the $n$-mode GDS is embedded into the product manifold, we can represent the relation between all modes of a tensor in a unified design. We can also go further and evaluate each mode separately, providing information to create a flexible measure of similarity, which is developed as the weighted geodesic distance. In summary, the main contributions of this paper are:

\begin{enumerate}[(i)]
    \item A novel tensor data representation called $n$-mode GDS.
    \item Integration of the $n$-mode GDS projection into the conventional product manifold, providing a tensor classification framework.
    \item Optimization of the proposed $n$-mode GDS projection on the product manifold space through a redefined Fisher score designed for tensor data.
    \item Introduction of an improved version of the geodesic distance, which incorporates the importance of each tensor mode for classification.
\end{enumerate}

We emphasize that one of the advantages of employing the PGM is that it allows the construction of a space comprising any combination of Grassmann manifolds~\citep{prodm_p}. In machine learning, this may simplify feature extraction and optimization since it is a differentiable manifold~\citep{prodm_1,prodm_2}. As pointed out by~\cite{exam_geo1}, the Grassmann manifold is a curved space, requiring a non-Euclidean distance to compute the similarity between elements. In practical terms, if the Euclidean distance is chosen to measure the similarity between elements on a Grassmann manifold, it can select element neighbors, which are actually distant if we consider the Grassman geometry~\citep{exam_geo}.

We have evaluated the proposed approach on five video datasets containing human actions and compared the results with the results achieved by other state-of-the-art approaches. The experimental results showed that $n$-mode GDS outperforms conventional subspace-based methods on action recognition in terms of accuracy. Moreover, the proposed $n$-mode GDS does not require pre-training, which is an advantage in several applications where pre-trained models are not available.

This paper is organized as follows. Section~\ref{S:Proposed} introduces the proposed method. The experimental results are provided in Section~\ref{S:Experimental}. Finally, Section~\ref{S:Conclusion} summarizes the main conclusions of this study and suggests possible future work.

\section{Proposed Method}
\label{S:Proposed}

In this section, we  first introduce the tensor matching problem, and then use this formulation to show the procedure for extracting subspaces from tensor data. We then present the GDS projection for providing discriminative properties. After that, we describe the formulation that encloses the geodesic distance and its improved version to compute the similarity between tensors. Finally, we present the Fisher score for $n$-mode subspaces. We use the following notation in this paper. Scalars are denoted by lower case letters and sets are denoted by boldface uppercase letters. Calligraphic letters are assigned to tensors. Given a matrix $A \in \mathbb{R}^{w \times h}$, $ A^{\top} \in \mathbb{R}^{h \times w}$ denotes its transpose.

\subsection{Problem Formulation}
Multi-dimensional data are usually represented by a set of modes ($n$-mode tensor) in order to reduce computational complexity. This procedure has the immediate advantage of allowing parallel processing. Furthermore, the $n$-mode tensor representation permits the examination of correlations among the various factors inherent in each mode.

Given two $n$-mode tensors $\mathcal A$ and $\mathcal B$, we can formulate the tensor matching problem in two steps. First, we create a convenient representation, where $\mathcal A$ and $\mathcal B$ can be expressed in a compact and informative manner. Second, we establish a mechanism for producing a reliable measure of similarity between these representations, allowing the comparison of $\mathcal A$ and $\mathcal B$.

\begin{figure*}[ht!]
    \centering
    \begin{subfigure}[]{0.3\textwidth}
        \scalebox{.6}{%

\begin{tikzpicture}

    
    
    \begin{axis}[colormap = {whitegray}{color=(white) color=(gray)}, view={-70}{30}, faceted color=gray, zmin=-0.3, zmax=1, mesh/ordering=y varies, mesh/cols=15, xticklabels={,,}, yticklabels={,,}, zticklabels={,,}]
    
    \addplot3[surf,z buffer=sort, restrict x to domain={-2:0}] table[ x index=0, y index=1, z index=2 ] {data.dat}; 
    \addplot3[surf, z buffer=sort, restrict x to domain={-4:-2}] table[ x index=0, y index=1, z index=2 ] {data.dat};
    \addplot3[surf, z buffer=sort, restrict x to domain={-8:-4}] table[ x index=0, y index=1, z index=2 ] {data.dat};

    
    \addplot3[black,mark=*, opacity =.9] coordinates {(-2, -3, -0.124111881387496)};
    \addplot3[black,mark=*, opacity =.9] coordinates {(-2, 3, -0.124111881387496)};

    \addplot3[mark=none, <->, red, line width=1pt] coordinates {(-2, -3, -0.124111881387496) (-2, 3, -0.124111881387496)};

    
    \node (A) at (axis cs:-2,  3.2, 0.07588811861) {\Large $\mathcal{A}$};
    \node (B) at (axis cs:-2, -3.2, 0.07588811861) {\Large $\mathcal{B}$};
    
    \end{axis}
    
\end{tikzpicture}}%
        \caption{Euclidean distance}
    \end{subfigure} \quad
    \begin{subfigure}[]{0.3\textwidth}
        \scalebox{.6}{%

\begin{tikzpicture}

    
    
    \begin{axis}[colormap = {whitegray}{color=(white) color=(gray)}, view={-70}{30}, faceted color=gray, zmin=-0.3, zmax=1, mesh/ordering=y varies, mesh/cols=15, xticklabels={,,}, yticklabels={,,}, zticklabels={,,}]
    
    \addplot3[surf,z buffer=sort, restrict x to domain={-2:0}] table[ x index=0, y index=1, z index=2 ] {data.dat}; 
    \addplot3[surf, z buffer=sort, restrict x to domain={-4:-2}] table[ x index=0, y index=1, z index=2 ] {data.dat};
    \addplot3[surf, z buffer=sort, restrict x to domain={-8:-4}] table[ x index=0, y index=1, z index=2 ] {data.dat};

    \addplot3[black,mark=*, opacity =.9] coordinates {(-2, -3, -0.124111881387496)};
    \addplot3[black,mark=*, opacity =.9] coordinates {(-2, 3, -0.124111881387496)};

    \addplot3[<->, restrict x to domain={-2:-2}, restrict y to domain={-3:3}, line width=1pt, black] table[ x index=0, y index=1, z index=2 ] {data.dat};

    \node (A) at (axis cs:-2,  3.2, 0.07588811861) {\Large $\mathcal{A}$};
    \node (B) at (axis cs:-2, -3.2, 0.07588811861) {\Large $\mathcal{B}$};
    
    \end{axis}
    
\end{tikzpicture}}%
        \caption{Geodesic distance}
    \end{subfigure} \quad
    \begin{subfigure}[]{0.3\textwidth}
        \scalebox{.6}{%

\begin{tikzpicture}

    
    
    \begin{axis}[colormap = {whitegray}{color=(white) color=(gray)}, view={-70}{30}, faceted color=gray, zmin=-0.3, zmax=1, mesh/ordering=y varies, mesh/cols=15, xticklabels={,,}, yticklabels={,,}, zticklabels={,,}]
    
    \addplot3[surf,z buffer=sort, restrict x to domain={-2:0}] table[ x index=0, y index=1, z index=2 ] {data.dat}; 
    \addplot3[surf, z buffer=sort, restrict x to domain={-4:-2}] table[ x index=0, y index=1, z index=2 ] {data.dat};
    \addplot3[surf, z buffer=sort, restrict x to domain={-8:-4}] table[ x index=0, y index=1, z index=2 ] {data.dat};

    \addplot3[black,mark=*, opacity =.9] coordinates {(-2, -6, 0.00653930773432966)};
    \addplot3[black,mark=*, opacity =.9] coordinates {(-2, 7, 0.115356136847076)};

    \addplot3[<->, restrict x to domain={-2:-2}, restrict y to domain={-6:7}, line width=1pt, blue] table[ x index=0, y index=1, z index=2 ] {data.dat};
    
    
    \node (A) at (axis cs:-2, 7, 0.36535613684) {\Large $\mathcal{{A}}$};
    \node (B) at (axis cs:-2, -6, 0.25653930773432966) {\Large $\mathcal{{B}}$};
    
    \end{axis}
    
\end{tikzpicture}}%
        \caption{Effect of $n$-mode GDS}
    \end{subfigure}
    \caption{Illustration of the geodesic and the Euclidean distance on the product manifold. (a) The Euclidean distance is the distance calculated when directly connecting $\mathcal A$ and $\mathcal B$, which is the shortest distance between them. (b) The geodesic distance uses the manifold surface, reflecting the actual distance between $\mathcal A$ and $\mathcal B$. (c) By employing the $n$-mode GDS projection, we improve the geodesic distance, since this reveals discriminative information.}\label{fig:geodesic}
\end{figure*}
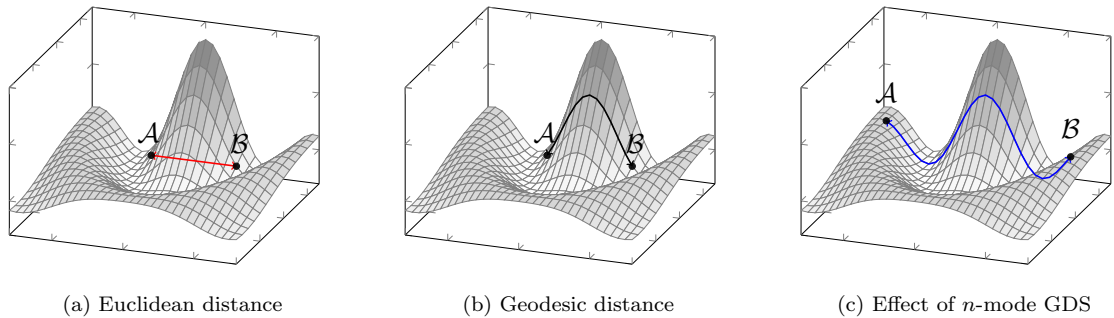

\subsection{Tensor Representation by Subspaces}
The tensors $\mathcal A$ and $\mathcal B$ represent distinct properties in each mode. For instance, in the case of video data where $n=3$, we have two spatial modes and a temporal one. Thus, each mode must be analyzed independently according to its factors. To simplify this procedure, we employ the unfolding process. We denote the set of unfolded images corresponding to the mode-$1$, mode-$2$, and mode-$3$ unfolding of $\mathcal A$ by $\bm X = \{X_{i}\}_{i=1}^n$. We perform the same procedure on the tensor $\mathcal B$, which gives $\bm Y = \{Y_{i}\}_{i=1}^n$.

Eigen-decomposition can be used to derive a set of eigenvectors for each element of $\bm X$ and $\bm Y$. It is expected that the eigenvectors associated with the largest eigenvalues of each element of $\bm X$ and $\bm Y$ accurately represent their elements in terms of variance maximization~\citep{PCA}. After selecting these eigenvectors, we obtain the following sets $\bm{U}_X = \{U_{i}\}_{i=1}^n$ and $\bm{U}_Y = \{U_{i}\}_{i=1}^n$. Since each mode expresses a distinct factor, it is reasonable to expect that each set of eigenvectors has a different distribution and properties, requiring a disjoint analysis to represent them accurately. For example, some modes may require more eigenvectors for their representation than others.

Now that we have $\bm{U}_X$ and $\bm{U}_Y$, which span the $n$-mode subspaces $\bm P = \{P_{i}\}_{i=1}^n$ and $\bm Q = \{Q_{i}\}_{i=1}^n$, we can employ a mechanism to extract more discriminative information from $\mathcal A$ and $\mathcal B$. According to the literature, subspaces are a very efficient tool for representing small and medium sized datasets. However, this representation may not be ideal for classification, since the subspaces are calculated independently without learning correlations between the distinct class distributions. At this point, we may employ some of the available techniques for enhancing the subspace representation~\citep{GDS,OS,Whitening,GODS} to create a set of subspaces $\bm D = \{D_{i}\}_{i=1}^n$, whereby projecting the sets $\bm P$ and $\bm Q$, we obtain suitable subspaces for classification. In our investigation, we adopt GDS~\citep{GDS} since it provides a reasonable balance between robustness and computational complexity considering that it is mainly based on eigen-decomposition. Once we have projected the $n$-mode subspaces $\bm P$ and $\bm Q$ onto $\bm D$, we obtain the sets $\hat{\bm P}$ and $\hat{\bm Q}$. After we select the similarity function, we have the essential components for representing and measuring the similarity between $\mathcal A$ and $\mathcal B$. The following sections present the details for computing $\bm D$.

\subsection{Generating $n$-mode Subspaces}
In order to compute the $n$-mode subspaces from tensor data, we employ the $n$-mode singular value decomposition ($n$-mode SVD)~\citep{MODE,HOSVD1}. The $n$-mode SVD provides a means for extracting basis vectors from unfolded tensors through the use of the traditional SVD. Given an $n$-mode tensor $\mathcal A$, the objective of $n$-mode SVD is to derive a set of orthogonal basis vectors $\bm{U} = \{U_{i}\}_{i=1}^n$ and a core tensor $\mathcal S$. By using $n$-mode SVD, every such tensor can be decomposed as follows:

\begin{equation}
\label{eq:product}
\mathcal{A} = \mathcal{S} \times U_{1} \times U_{2} \times \ldots \times U_{n},
\end{equation}

\noindent where the core tensor $\mathcal S$ includes information about the various mode matrices $U_{1}, U_{2}, \ldots, U_{n}$, and each mode matrix $U_{i}$ contains the orthonormal vectors spanning the column space of the matrix $X_{i}$, which is the result of tensor $\mathcal A$ unfolding. This decomposition provides flexibility since it gives the tools for analyzing each tensor factor independently. By employing this strategy, we also preserve the computational complexity of SVD since $n$-mode SVD can be implemented by a series of $n$ SVD decompositions. It is important to note that the employed collection of matrices $\bm{X}$ do not satisfy the zero expectation condition, (i.e., $E(X_{i}) \neq 0$, $\forall \ X_{i} \in \bm{X}$), contrasting with the originally proposed $n$-mode SVD~\citep{MODE,HOSVD1}.

Previous studies indicate that $\mathcal S$ contains rich information regarding the relation between the set $\bm U$ and can be used in classification and reconstructions methods~\citep{LIT1,LIT2}. In spite of its importance, we employ the average canonical angle to describe the relation between $n$-mode subspaces, which does not require $\mathcal S$.

\subsection{Selecting the $n$-mode Subspace Dimensions}
One of the benefits of using subspaces to represent tensors is that it provides a useful mechanism for defining the compactness ratio, that is, how much information of the patterns of a particular mode should be preserved to maintain the trade-off between data contribution concerning variance and subspace dimension. Given one tensor-unfolded mode $X$, Equation~\eqref{eq:dimension} defines the proportion of the basis vectors employed to describe $X$ compactly~\citep{CITE1,CITE2}:

\begin{equation} \label{eq:dimension}
\mu(K) \leq 100\% \times \frac{\sum_{k=1}^{K}(\lambda_k)}{\sum_{k=1}^{R}(\lambda_k)}.
\end{equation}

\noindent where $K$ is the number of selected basis vectors that span a subspace $P$, $\lambda_k$ is the $k$-th eigenvalue of $X$, and $R=\mathrm{rank}(X)$. The function $\mu(\cdot)$ controls the trade-off between the compactness ratio of $X$ and the amount of accumulated energy in the first $k$ eigenvectors. This parameter depends on the complexity of the linear correlations of each tensor mode and is also application-dependent. For example, when we have a high correlation between the matrices of a particular mode or there are repeated exemplars, the subspace representation exhibits a compact shape; that is, the first eigenvectors associated with the first eigenvalues explain most of the data.

\subsection{Generating the $n$-mode GDS Projection}
In an $m$-class classification problem, $\bm P=\{P_{ij}\}_{i,j=1}^{n,m}$ denotes the set of all $n$-mode subspaces spanned by $\bm U = \{U_{ij}\}_{i,j=1}^{n,m}$. We can then develop the $n$-mode GDS projection $\bm D = \{D_{i}\}_{i=1}^n$ that act on $\bm P$ to extract discriminative information. Since each mode subspace reflects a particular factor, it is essential to handle each one independently and compute a model that reveals hidden discriminative structures. In traditional GDS, this procedure is performed through the removal of overlapping components that represent the intersection between subspaces. In mathematical terms, the GDS projection can be described as the extension of the difference vector between two vectors in a multi-dimensional space.

By discarding the components that express the intersection between subspaces, GDS consists entirely of the required elements for classification~\citep{GDS,GODS}. Therefore, by projecting the subspaces onto the $n$-mode GDS, we expect to extract suitable information for tensor classification. Figure~\ref{fig:geodesic} shows the advantages of using the $n$-mode GDS projection on PGM. Whereas the Euclidean distance neglects the manifold surface, which may result in information loss, the geodesic distance uses the manifold surface, and the $n$-mode GDS projection improves the distance between the different $n$-mode subspace classes. In order to compute the $n$-mode GDS, we compute the sum of the projection matrices of each $i$-mode subspace as follows:

\begin{equation} \label{eq:sum_subspace}
G_{i}=\frac{1}{m}\sum_{j=1}^m U_{ij} U_{ij}^{\top}, \quad \textrm{for} \ 1 \leq i \leq n.
\end{equation}

Since $G_{i}$ has information regarding all class subspaces in a particular mode, it is beneficial to decompose it to exploit discriminative elements. Applying eigen-decomposition to $G_{i}$, we obtain

\begin{equation} \label{eq:GDS}
G_{i} = V_{i} \Sigma_{i} V_{i}^{\top}, \quad \textrm{for} \ 1 \leq i \leq n,
\end{equation}

\noindent where the columns in $V_i = \{\phi_1, \phi_2, \ldots, \phi_{R_i}\}$ are the normalized eigenvectors of $G_{i}$, and $\Sigma_{i}$ is the diagonal matrix with corresponding eigenvalues $\{\lambda_1, \lambda_2, \ldots, \lambda_{R_i}\}$ in descending order, where $R_i = \mathrm{rank}(G_{i})$. The $n$-mode GDS projection discards the first few eigenvectors of $G_{i}$ with large eigenvalues and retains only the last few eigenvectors of $G_{i}$ with small eigenvalues. Thus, the $n$-mode GDS provides the difference information between $n$-mode class subspaces. Therefore, we can define $D_{i} = \{\phi_{\alpha_i}, \ldots, \phi_{\beta_i} \}$, where $\alpha_i < \beta_i \leq R_i$. The $n$-mode GDS dimension is defined by maximizing the mean canonical angles between $n$-mode class subspaces.

\begin{figure*}[ht!]
	\label{fig:framework}
	\centering
	\def\svgwidth{\linewidth}         
	\fontsize{8pt}{11pt}\selectfont   
	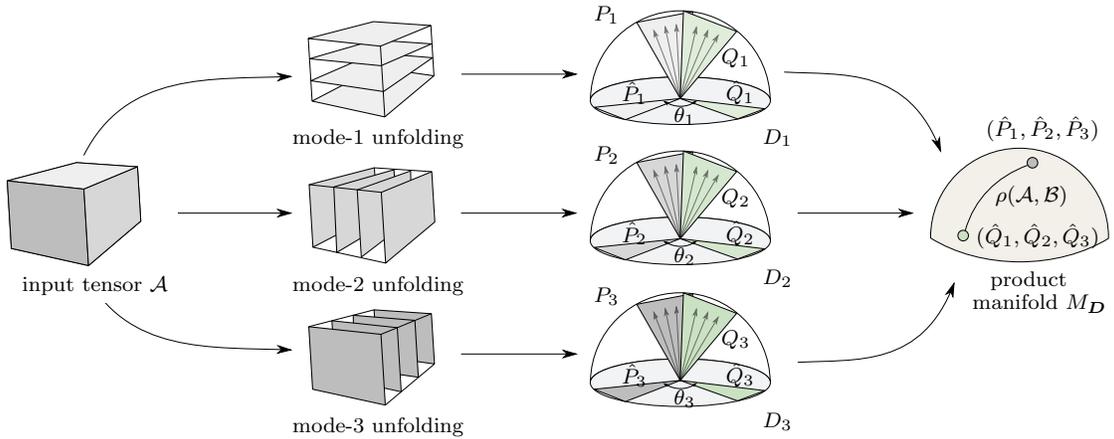
	\caption{Conceptual figure of the $n$-mode GDS projection. First, we unfold the $3-$mode tensor $\mathcal{A}$ and compute its subspaces from the unfolded modes. Next, we project the subspaces onto the $n$-mode GDS. The product of manifolds can be used to represent the projected subspaces. Finally, the chordal distance $\rho(\mathcal{A}, \mathcal{B})$ determines the similarity between tensors $\mathcal{A}$ and $\mathcal{B}$.}
\end{figure*}

\subsection{Projecting the $n$-mode Subspaces onto the $n$-mode GDS}

Once $\bm D = \{D_i\}_{i=1}^n$ is computed, we can extract more discriminative structures from $\bm P = \{P_{ij}\}_{i,j=1}^{n,m}$. According to \citet{CMSM} and \citet{GDS}, this can be achieved by two different approaches. The first approach involves projecting subspaces onto a discriminative space, then orthogonalizing the projected subspaces using Gram-Schmidt orthogonalization. The second procedure includes projecting subspaces onto a discriminative space directly, then applying SVD to generate the projected subspaces. \citet{CMSM} and \citet{GDS} established that these two procedures are algebraically equivalent. In this work, we employ the first procedure, which is consistent with the conventional method. Therefore, the procedure to compute $\hat{\bm P} = \{\hat{P}_{ij}\}_{i,j=1}^{n,m}$ is

\begin{equation} \label{eq:project}
\hat{P}_{ij} = \mathrm{orth}\left(D_{i}^{\top} P_{ij}\right),
\end{equation}

\noindent where the $\mathrm{orth}(\cdot)$ operator denotes the orthogonalization and normalization of a set of vectors using Gram-Schmidt orthogonalization.

\subsection{Representing the $n$-mode Subspaces $\hat{\bm P}$ on the Product Manifold}
We introduce the product manifold to describe $\hat{\bm P}$ as a single manifold. This manifold consists of the product of the projected $n$-mode subspaces onto the $n$-mode GDS. In traditional PGM, however, the subspaces are generated directly from the tensors by employing $n$-mode SVD. Although this procedure provides a convenient representation for the tensors, the generated subspaces may not be ideal for classification. In contrast, we project ${\bm P}$ onto $\bm D$ before applying the product manifold. Our objective is to achieve more efficient subspaces for classification. Therefore, given a set of manifolds $\bm {M} = \{M_{i}\}_{i=1}^n$ composed by $\hat{\bm P}$, Equation~\eqref{eq:PM} describes the product manifold:

\begin{equation} \label{eq:PM}
M_{\bm D} =  M_{1} \times M_{2} \times \ldots \times M_{n} = \big(\hat{P}_{1}, \hat{P}_{2}, \ldots, \hat{P}_{n}\big),
\end{equation}

\noindent where $\times$ denotes the Cartesian product, $M_{i}$ is a $i$-mode manifold, and $\hat{P}_{i} \in M_{i}$. It is worth noting that the manifold topology of $M_{\bm D}$ is equivalent to the product topology~\citep{prod1}. The advantage of using $M_{\bm D}$ is that it provides a combined topological space associated with $\hat{\bm P}$. For illustration, in gesture and action recognition problems, where we handle $3$-mode tensors, we can replace the tensor representation by elements on a product manifold. Therefore, tensor data can be regarded as points on the product manifold $M_{\bm D}$. Another benefit of employing $M_{\bm D}$ to represent tensor data is that it provides the means for working directly with geodesics through the use of the geodesic distance. The geodesic distance between two points is the length of the geodesic path, which is the shortest path between the points that lie on the surface of the manifold. Furthermore, the geodesic distance provides a more accurate similarity between two points on the product manifold, since it uses the surface of the manifold ~\citep{GEO_REASON}. Figure~\ref{fig:geodesic} illustrates the geodesic distance.

\subsection{Fisher score for $n$-mode Subspaces}
We now introduce the Fisher score for tensorial class separability index. Traditionally, the Fisher score $F(\Psi)$ of a transformation matrix $\Psi$ can be defined as the ratio of two variables: $F(\Psi) = F^{\,b}/F^{\,w}$, where $F^{\,b}$ and $F^{\,w}$ are the inter-class and intra-class variability, respectively. Therefore, a high Fisher score ensures high inter-class and low intra-class variability. We extend the Fisher score to evaluate subspace separability by re-defining the $F^{\,b}$ and $F^{\,w}$ scores as follows:



\begin{align}
  F^{\,b} ={}& \frac{1}{m} \sum_{j=1}^{m} \mathrm{Sim}\big( K_j,~K \big). \\
  F^{\,w} ={}& \frac{1}{v} \sum_{j=1}^{m} \sum_{l=1}^{m_j} \mathrm{Sim}\big( K_{jl},~K_j \big).
\end{align}

In Equations~(7) and (8), $m$ is the number of tensor classes, where each $j$-th class represents $m_j$ tensors, $K_j$ represents the Karcher mean of the $j$-th class subspace, $K$ is the Karcher mean of the $m$ $K_j$ subspaces, and $v = \sum_{j=1}^{m} m_j$, where $m_j$ is the number of subspaces in the $j$-th class.

Finally, $\mathrm{Sim}(\cdot,\cdot)$ is a function that measures the similarity between subspaces using the geodesic distance. $\mathrm{Sim}(\cdot,\cdot)$ should explore the manifold surface, considering that we are dealing with subspaces. Additionally, the geodesic variants of MDS and CCA have shown to be effective in dimensionality reduction and classification, outperforming their Euclidean equivalents in many scenarios~\citep{geodesic_2,geodesic_1}.

We can now aggregate the inter-class and intra-class variability from the $n$-mode subspaces by summing up $F^{\,b}$ and $F^{\,w}$ from their respective modes using Equations (7) and (8). Since $F(\cdot)$ is not defined to handle $n$-mode subspaces, we adapt the Fisher score to the $n$-mode case as the average of the Fisher scores in each mode, as follows:

\begin{align}
  F_n^{\,b} ={}& \frac{1}{n} \sum_{i=1}^{n} F_{i}^{\,b}, \\
  F_n^{\,w} ={}& \frac{1}{n} \sum_{i=1}^{n} F_{i}^{\,w}.
\end{align}

The $F_{n}(\bm \Phi) = F_n^{\,b}/F_n^{\,w}$ score is the class separability index for $n$-mode subspaces, where $\bm \Phi = \{\Phi_i\}_{i=1}^{n}$ is a set of transformation matrices. The introduced $n$-mode Fisher score is used as an evaluation tool for selecting the optimal dimension of $\bm D$. 

\subsection{Weighted Geodesic Distance}
We employ the average of the canonical angles to compare the subspaces of different modes. A practical technique for computing the canonical angles between two subspaces $P$ and $Q$ is by computing the eigenvalues of the product of their basis vectors. Therefore, given $U_{P}$ and $U_{Q}$, which span the subspaces $P$ and $Q$, Equation~\eqref{eq:angles} computes the canonical correlations between $P$ and $Q$.

\begin{equation}
\label{eq:angles}
U_{P}^{\top} U_{Q} = U \Sigma V^{\top}.
\end{equation}

\noindent where the eigenvalues matrix $\Sigma$ provides the canonical correlations between the principal angles of $U_{P}$ and $U_{Q}$ and can be used to compute the canonical angles, since $\Sigma = \mathrm{diag}(\lambda_1, \lambda_2, \ldots, \lambda_{K})$. The canonical angles $\{\theta_k\}_{k=1}^{K}$ can then be computed by using the inverse cosine of $\Sigma$, as $\{\theta_k=\cos^{-1}(\lambda_k) \}_{k=1}^{K}$. Finally, the average canonical angle $\bar\theta$ between $P$ and $Q$ is defined as $\bar\theta = \frac{1}{K} \sum_{k=1}^{K} \theta_k$, where $K \leq \mathrm{min}(\mathrm{rank}(U_{P}), \mathrm{rank}(U_{Q}))$. Once the average of the canonical angles is obtained in all available modes $\bar{\bm \theta} = \{\bar\theta_i\}_{i=1}^{n}$, we can introduce the weighted geodetic distance based on the product manifold, which is defined as

\begin{equation}
\label{eq:frob}
\rho(\mathcal A, \mathcal B) = \left(\sum_{i=1}^{n}\left(w_i\bar\theta_i\right)^2\right)^{1/2},
\end{equation}

\noindent where we estimate $w_i$ by using the Fisher score since each mode gives a different separability index reflecting the importance of each mode in terms of classification:

\begin{equation}
\label{eq:weight}
w_i =  \frac{ F(D_i) } {\sum_{i=1}^{n}F(D_i)},
\end{equation}

\noindent where $F(D_i)$ is the Fisher score for the projection matrix $D_i$. It is worth noting that when $w_i = 1$, this distance on the Cartesian product is regarded as the product manifold distance. The geodesics in the product manifold $M_D$ are just the products of geodesics in the factor manifolds $\bm {M} = \{M_{i}\}_{i=1}^n$. By employing $w_i$, we can exploit the importance of each factor manifold, which can improve the classification accuracy.

In Equation~\eqref{eq:frob}, we need to minimize $\rho(\cdot,\cdot)$ when tensors $\mathcal A$ and $\mathcal B$ are observations of the same class. Otherwise, when $\mathcal A$ and $\mathcal B$ represent distinct classes, $\rho(\cdot,\cdot)$ needs to return a high value. Since $\rho(\cdot,\cdot)$ is essentially proportional to the average of the canonical angles between $\hat{\bm P}$ and $\hat{\bm Q}$, its optimization process requires only the proper selection of $\alpha_i$, $\beta_i$ and $w_i$. We can achieve quasi-orthogonality between the $n$-mode subspaces by generating the appropriate GDS projection $\bm D$, thereby enlarging its geodesic distance on the product manifold. Formally, we can obtain $\bm D$ as follows:

\begin{equation} \label{eq:optimization}
\bm D = \argmax F_n(\bm V^\prime),
\end{equation}


\noindent where $\bm V^\prime = \{V_{i}^\prime\}_{i=1}^{n}$ and $V_{i}^\prime = \{\phi_{\alpha_i}, \ldots, \phi_{\beta_i}\}$ is the eigenvector subset of $\bm V$ obtained by Equation~\eqref{eq:GDS}. In the next section, we provide experimental results that support our claim.

\section{Experimental Results}
\label{S:Experimental}

We next evaluate the $n$-mode GDS projection to show its advantages over tensor-based methods for action and gesture recognition problems. First, we present the datasets and the experimental protocol employed. Next, we provide a visualization of the difference between the $n$-mode subspaces. After that, we evaluate the model discriminability by using the $n$-mode Fisher score. We then compare the proposed approach with related methods. The tensor modes are evaluated independently and simultaneously, followed by comparison with related methods. Finally, feature extraction techniques are employed on an $n$-mode GDS projection framework and a comparison with the state-of-the-art is presented.

\subsection{Datasets and Experimental Protocol}

Our experiments were conducted using five datasets. KTH~\citep{KTH} is a video dataset containing six types of human actions (walking, jogging, running, boxing, hand waving, and hand clapping) executed by $25$ subjects in four different scenarios: outdoors, outdoors with scale variation, outdoors with different clothes, and indoors. This dataset consists of $2~391$ sequences, where all videos were recorded over homogeneous backgrounds with a static camera (in most sequences, but hard shadows are present) with a temporal resolution of $25$ frames per second. In addition, there are significant variations in length and viewpoint. The videos were down sampled to 160$\times$120 pixels and have a length of four seconds on average. The videos are divided with respect to the subjects into a training set ($8$ persons), a validation set ($8$ persons), and a test set ($9$ persons).

We employ the Cambridge Gesture dataset~\citep{kim_dataset} containing $900$ video sequences of nine hand gesture classes. Each gesture class consists of $100$ video sequences, with the videos collected from five different illumination sets designated as set1, set2, set3, set4, and set5. The set5 is used for the training and the other four sets are employed as test sets.

The UCF-101~\citep{action101} dataset is a large action recognition dataset that comprises $13~320$ YouTube video clips of $101$ action categories, separated into five categories: Human-Object Interaction, Body-Motion Only, Human-Human Interaction, Playing Musical Instruments, and Sports. Most of the videos are related to actions performed in sports. The videos duration varies from 2 to 15 seconds, with 25 frames per second.

The HMDB-51~\citep{action51} dataset contains $6~766$ video clips, where $3~570$ are employed for training and $1~530$ for testing. This dataset provides $51$ classes that were obtained from multiple sources, including movies, YouTube, and Google. Both UCF-101 and HMDB-51 datasets provide $3$ (training, testing)-folds, and we report the average accuracy of the three testing splits. Due to their complexity, in our experiments, we resize UCF-101 and HMDB-51 videos to 340$\times$256. Compared with UCF-101, the videos in HMDB-51 are more difficult since they present the complexity of real-world actions. The performance of these two datasets is calculated using the average accuracy.

The Osaka University Kinect Action Dataset~\citep{Osaka} contains $10$ actions performed by eight subjects collected by Osaka University. Action types consist of jumping jack type 1, jumping jack type 2, jumping on both legs, jumping on the right leg, jumping on the left leg, running, walking, side jumps , skipping on the left leg, and skipping on the right leg. The videos were down sampled from 320$\times$240 to 160$\times$120 pixels and have a length varying from $2$ to $4$ seconds. In addition to the RGB data, this dataset provides depth and skeleton data. During the data recording process, there were some variations in the illumination conditions and background. In our experiments, we employed the depth information to segment the foreground from the background pixels. For evaluation purposes, we use the leave-one-out cross-validation scheme.



\subsection{Visualization of the $n$-mode GDS Projection}

In this experiment, we aim to show the visual differences between the images of the unfolded tensors, the basis vectors, the $n$-mode sum subspace, and the $n$-mode principal subspace. The $n$-mode GDS projection is computed using two classes of the KTH dataset (boxing and waving).

\begin{figure*}[htpb]
    \center\includegraphics[width=\textwidth]{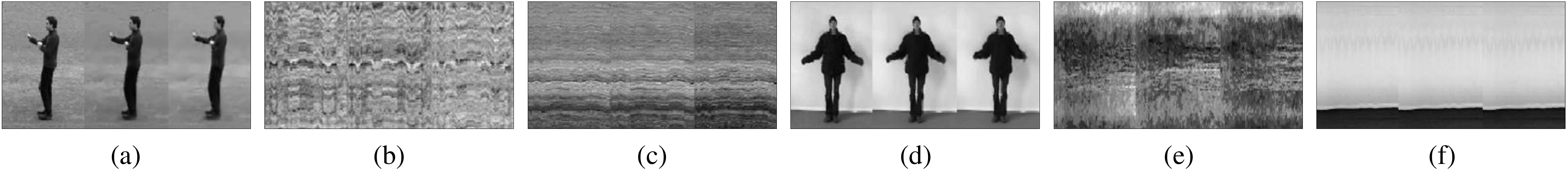}
    \caption{\label{fig:unfolded} Unfolded tensors of the boxing and waving classes of the KTH dataset.}
\end{figure*}

\begin{figure*}[htpb]
    \center\includegraphics[width=\textwidth]{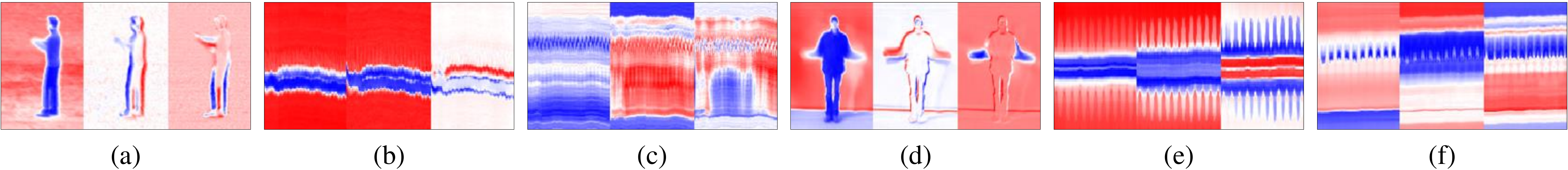}
    \caption{\label{fig:sum_subspace} Basis vectors of the unfolded tensors.}
\end{figure*}

\begin{figure*}[htpb]
    \center\includegraphics[width=\textwidth]{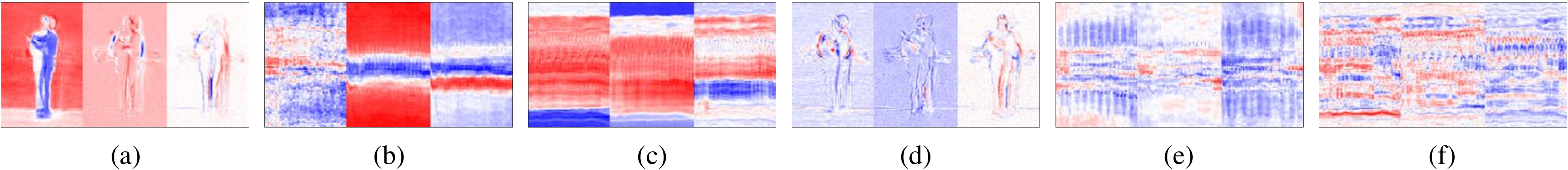}
    \caption{\label{fig:diff_subspace} The $n$-mode principal space and the $n$-mode GDS basis vectors.}
\end{figure*}

Figure~\ref{fig:unfolded} shows images of the unfolded tensors. Figures~\ref{fig:unfolded}(a),~\ref{fig:unfolded}(b) and~\ref{fig:unfolded}(c) show the first $3$ frames of the $1$-, $2$- and $3$-mode unfolding of the boxing class. Figures~\ref{fig:unfolded}(d),~\ref{fig:unfolded}(e) and~\ref{fig:unfolded}(f) show the first $3$ frames of the $1$-, $2$- and $3$-mode unfolding of the waving class. Figure~\ref{fig:sum_subspace} shows the first $3$ basis vectors of the respective unfolded tensors. Finally, Figure~\ref{fig:diff_subspace} shows the principal subspace and the difference subspace, which are the decomposition results of the sum subspace. Figures~\ref{fig:diff_subspace}(a),~\ref{fig:diff_subspace}(b) and~\ref{fig:diff_subspace}(c) show the common structures contained in both unfolded tensors. These structures provide low discriminative information since projecting different class subspaces results in closer subspaces and thus decreases the canonical angles between them. Figures~\ref{fig:diff_subspace}(d),~\ref{fig:diff_subspace}(e) and~\ref{fig:diff_subspace}(f) show the difference between the components contained in both unfolded tensors. In mathematical terms, the $n$-mode GDS projection represents the linear combination of the difference between the samples of a particular tensor mode. This linear combination preserves the discriminative structures in the form of a subspace, where projecting similar subspaces representing different classes results in enlarged canonical angles.

\subsection{Evaluating the $n$-mode GDS Projection Separability Using the $n$-mode Fisher Score}

We next evaluate the separability of the $n$-mode subspaces using both multi dimensional scaling (MDS) and the proposed $n$-mode Fisher score on the Osaka dataset. MDS enables the visualization of the similarity between $n$-mode subspaces by projecting the pairwise canonical angles among the subspaces onto an abstract space. In this experiment, the video data from the Osaka Kinect dataset was pre-processed by using a people detector, and the detected patches were cropped. Considering that the cropped patches have different sizes, we also normalized them to 30$\times$90. Linear interpolation was then employed to normalize the number of frames to compound the 30$\times$90$\times$30 tensors. We denote the modes obtained by tensor unfolding as follows: $1$-mode denotes the unfolding in the temporal $\bm t$-axis direction, $2$-mode represents the unfolding of the tensor in the spatial $\bm y$-axis direction, and the $3$-mode is described by unfolding in the spatial $\bm x$-axis direction.





The proposed $n$-mode GDS projection is compared with PGM obtained using MSM on the unfolded modes. When we compare MSM to the proposed method, we observe that, even though MSM may operate directly on tensors, it works with only one of the modes. Thus, the relation between MSM, GDS, PGM, and $n$-mode GDS is as follows. MSM executes the pattern-set matching using only one of the available tensor modes. GDS executes the same strategy as MSM, however, employing a discriminative space to improve its classification accuracy. PGM, in turn, applies the same approach as MSM but operates in all available tensor modes. Lastly, $n$-mode GDS utilizes a discriminative space produced by GDS, but in each separate mode and executes the fusion of those subspaces through the product of manifolds. Therefore, this experiment evaluates whether it is worth using the three modes, a specific combination (e.g., $1$-mode and $3$-mode), or only one of the modes.



 The accuracy of MSM was maximized when the dimensions of the subspaces were set to $15$, $10$, and $12$ for the $1-$, $2-$, and $3-$mode unfolding, respectively. The number of canonical angles that results in the best accuracy is $5$ for the $2-$ and $3-$mode unfolding and $7$ for the $1-$mode unfolding. Using a $10-$fold cross-validation scheme, MSM obtained a reasonable accuracy of $74.30\%$ on the $1-$mode unfolding, followed by $71.60\%$ and $62.90\%$ on the $2-$ and $3-$mode unfolding, respectively. By using the same set of parameters, PGM achieved an accuracy of $77.67\%$. Accordingly, the number of basis vectors required to represent the videos is very low compared to the original amount of data contained in the dataset. Moreover, the accuracy attained by MSM on each separated mode supports the idea that the modes should be used in a unified framework and a discriminative mechanism should be employed.



\begin{table}[]
\centering
\small
\caption{The accuracy and the $n$-mode Fisher score (in parenthesis) for the MSM and GDS subspaces.}
\label{tab:tab_fisher}
\begin{tabular}{@{}lccc@{}}
\toprule
\multicolumn{1}{c}{method} & 1-mode         & 2-mode         & 3-mode         \\ \midrule
MSM                        & 74.30\% (0.57) & 71.60\% (0.41) & 62.90\% (0.46) \\
GDS                        & 81.10\% (0.62) & 76.50\% (0.49) & 65.20\% (0.51) \\ \bottomrule
\end{tabular}
\end{table}

\begin{figure*}[htpb!]
    \center\includegraphics[width=\textwidth]{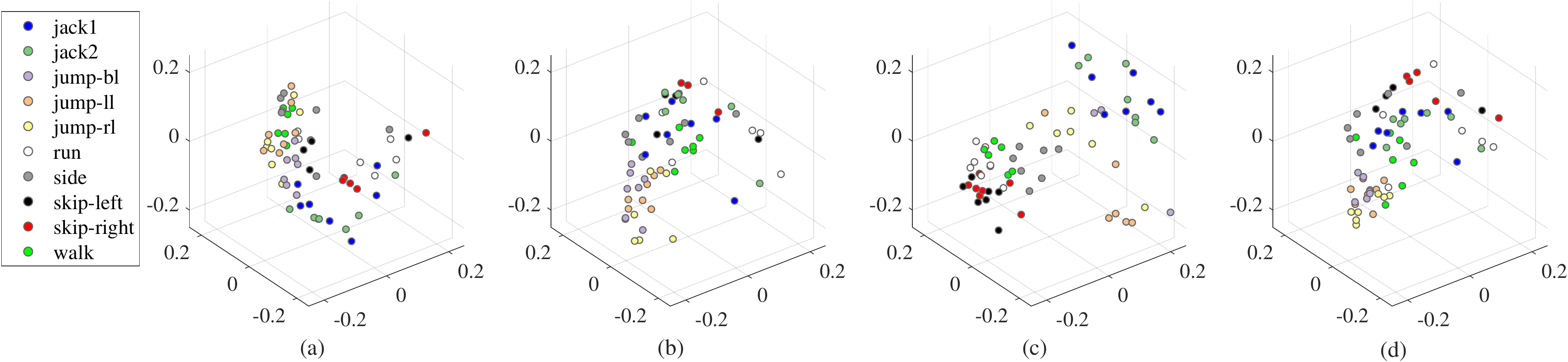}
    \caption{\label{fig:MDS_PGM} 3D scatter plots of Osaka Kinect dataset using MSM. (a) $3-$mode, (b) $2-$mode, and (c) $1-$mode unfolding are represented using MSM. PGM is shown on (d).}
\end{figure*}

\begin{figure*}[htpb!]
    \center\includegraphics[width=\textwidth]{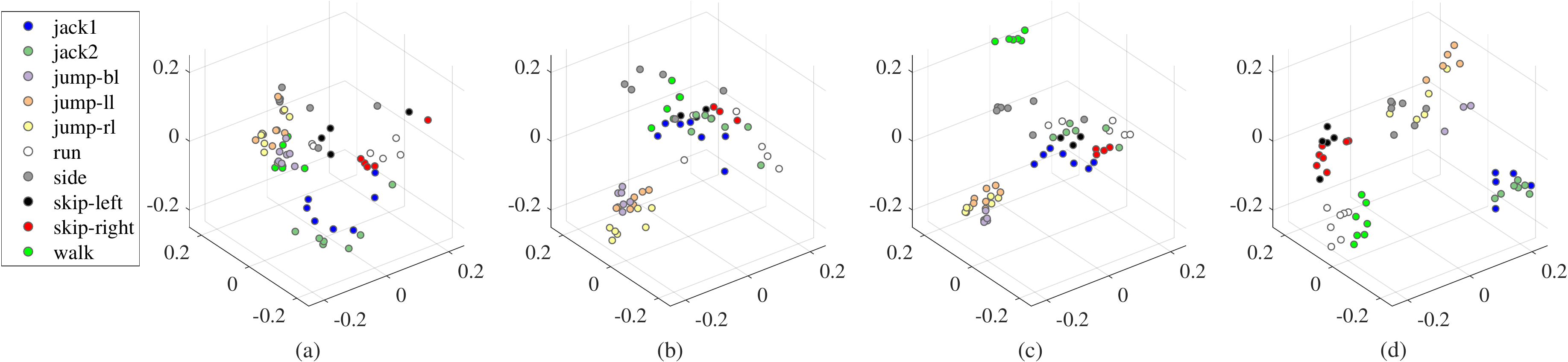}
    \caption{\label{fig:MDS_n-mode_GDS} 3D scatter plots of Osaka Kinect dataset using GDS. (a) $3-$mode, (b) $2-$mode and (c) $1-$mode unfolding are represented using GDS. $n$-mode GDS is shown on (d).}
\end{figure*}

Figure~\ref{fig:MDS_PGM} shows MDS plots of the subspaces obtained using MSM. This figure suggests that the distances in the subspaces change according to the mode unfolding, confirming that each mode may contribute differently to the classification accuracy. The $1-$mode unfolding shows more compact clusters than the others, suggesting that the $1-$mode provides slightly more robust features for classification. Figure~\ref{fig:MDS_PGM}(d) shows the arrangement of PGM. Compared with $1-$mode unfolding, PGM provides lower intra-class separability, although the clusters of the different classes appear closer, suggesting that the fusion schema adopted by PGM does not take into consideration the relationships between the different classes.

It is worth mentioning that PGM gives reasonable results in this dataset. Although the Osaka Kinect dataset has only ten classes, it provides a challenge for PGM since the inter-class distance of some classes are very low due to the similar actions contained in the dataset. For instance, the classes jumping with the right leg and jumping with the left leg present subspaces with a high amount of structural overlap, since just a small part of the subject body differentiates the classes. Taking into account that this discriminative element may not be present in the subspaces, MSM and PGM are not able to correctly represent these cases. 

The $n$-mode GDS projection exhibited the best accuracy when the dimensions of the subspaces were set to $17$, $12$, and $15$ for the $1$-, $2$- and $3$-mode unfolding, respectively. The number of canonical angles that resulted in the best accuracy was $5$ for all the unfolding modes. Using a $10$-fold cross-validation scheme, the $n$-mode GDS projection achieved an accuracy of $81.10\%$ on the $1$-mode unfolding, followed by $76.50\%$ and $65.20\%$ on the $2$- and $3$-mode unfolding respectively. By employing the same set of parameters, the $n$-mode GDS using all available modes achieved an accuracy of $83.30\%$.


Figure~\ref{fig:MDS_n-mode_GDS} shows MDS plots of the subspaces using the $n$-mode GDS projection. The clusters formed by the introduction of GDS on the unfolded tensors act by reducing the intra-class distance, creating more recognizable clusters. 
Figure~\ref{fig:MDS_n-mode_GDS}(d) shows the relation between the $n$-mode subspaces when discriminative structures are available. Visually, it is possible to observe that the proposed method provides a lower intra-class distance, while improving the inter-class distance. It is worth mentioning that the increase in inter-class distance seems to occur in all classes, although very similar classes (such as jumping on both legs, jumping on the right leg, and jumping on the left leg) still exhibit much overlap, which is expected. Since PGM has no mechanism to extract discriminative structures, it relies on the data distribution itself, depending on the structural differences between the tensor subspaces. When two $n$-mode subspaces representing different classes exhibit high overlap, the similarity between them is high, and the subspace accuracy decreases. However, $n$-mode GDS can detect and remove the common components existing in similar classes, exposing only the relevant components for classification. In algebraic terms, $n$-mode GDS enlarges the canonical angles between similar classes, since the common structures between the subspaces are removed. This observation is also supported by the computed $n$-mode Fisher score of MSM and GDS, which are listed in Table~\ref{tab:tab_fisher}.

\subsection{Evaluating Tensor Modes}

In the following experiments, we use the KTH and Cambridge gesture datasets. We evaluate the tensor modes independently and in combinations to determine their contribution to the recognition rate. In this experiment, we compare our proposed framework with MSM, GDS, and PGM. These methods employ the concept of subspaces and canonical angles, which may provide an objective interpretation of each mode subspace. In video data, since each mode assigns to a different factor, each subspace has a distinct contribution in the classification. In addition, as advocated in the literature of feature subset selection, two subsets of attributes which do not operate adequately independently may perform very well when employed in combination~\citep{feat_sel}.

Following the experimental protocol of~\citet{KIM_DCC2}, the video data in the KTH and Cambridge gesture datasets were resized to 20$\times$20$\times$20. The subspace dimension $K$ was chosen to encode $90\%$ of the variance in the original data. Tables~\ref{tab:tab_modes_cam} and ~\ref{tab:tab_modes_kth} list the accuracy results achieved by MSM and GDS on different modes using Cambridge and KTH datasets respectively. When $1$-mode unfolding is employed, both MSM and GDS achieved the highest scores. As expected, GDS outperformed MSM in both scenarios.

\begin{table}[htpb!]
\centering
\small
	\caption{Average accuracy and standard deviation of different modes and combinations using the Cambridge dataset.}
	\label{tab:tab_modes_cam}
\resizebox{0.7\columnwidth}{!}{
\begin{tabular}{@{}lcccc@{}}
\toprule
approach    & 1-mode               & 2-mode               & 3-mode               & $-$                  \\ \midrule 
MSM         & 64.55 $\pm$ 4.9      & 40.21 $\pm$ 5.9      & 56.29 $\pm$ 5.3      & $-$                  \\
GDS         & 78.29 $\pm$ 3.7      & 51.50 $\pm$ 4.3      & 67.24 $\pm$ 4.1      & $-$  \\ \midrule 
approach    & $\bm a$-mode               & $\bm b$-mode               & $\bm c$-mode               & $\bm d$-mode               \\ \midrule 
PGM~\citep{LUI2} & 68.35 $\pm$ 2.2 & 79.23 $\pm$ 2.2 & 61.67 $\pm$ 2.4 & 88.13 $\pm$ 2.1 \\
$n$-mode GDS     & 74.17 $\pm$ 2.2 & 88.76 $\pm$ 2.1 & 71.33 $\pm$ 2.3 & 93.51 $\pm$ 2.1 \\
$n$-mode wGDS    & 74.49 $\pm$ 2.1 & 89.11 $\pm$ 2.0 & 71.48 $\pm$ 2.3 & \textbf{94.25 $\pm$ 1.9} \\ \bottomrule
\end{tabular}}
\end{table}

\begin{table}[htpb!]
\centering
\small
	\caption{Average accuracy and standard deviation of different modes and combinations using the KTH dataset.}
	\label{tab:tab_modes_kth}
\resizebox{0.7\columnwidth}{!}{
\begin{tabular}{@{}lcccc@{}}
\toprule
approach    & 1-mode               & 2-mode               & 3-mode               & $-$                  \\ \midrule
MSM         & 83.03 $\pm$ 3.5      & 67.12 $\pm$ 4.3      & 71.37 $\pm$ 3.9      & $-$                  \\
GDS         & 91.51 $\pm$ 1.9      & 70.78 $\pm$ 1.5      & 83.45 $\pm$ 2.1      & $-$  \\ \midrule
approach     & $\bm a$-mode               & $\bm b$-mode               & $\bm c$-mode               & $\bm d$-mode                \\ \midrule
PGM~\citep{LUI2} & 80.15 $\pm$ 2.8 & 85.73 $\pm$ 2.6 & 74.56 $\pm$ 2.9 & 96.17 $\pm$ 1.7 \\
$n$-mode GDS     & 83.84 $\pm$ 2.1 & 91.28 $\pm$ 1.9 & 78.54 $\pm$ 2.1 & 97.33  $\pm$ 1.6 \\
$n$-mode wGDS    & 84.16 $\pm$ 2.1 & 91.67 $\pm$ 1.9 & 78.34 $\pm$ 2.1 & \textbf{98.64  $\pm$ 1.4} \\ \bottomrule
\end{tabular}}
\end{table}

Tables~\ref{tab:tab_modes_cam} and~\ref{tab:tab_modes_kth} also show the results of PGM, $n$-mode GDS, and $n$-mode wGDS (weighted GDS) when the modes are combined, where the $\bm a$-, $\bm b$-, $\bm c$-, and $\bm d$-mode are the combinations $1$-$2$, $1$-$3$, $2$-$3$, and $1$-$2$-$3$, respectively. The results show that mode combinations improve the accuracy of all methods, indicating that the time information is decisive discriminative information. In addition, the best results are obtained when all the available modes are employed. The weighted geodesic distance strategy was shown to be efficient, validating the strategy of using weights at each geodesic distance.


\subsection{Comparison with Related Methods}

For this comparison, we employ the following traditional methods: discriminative canonical correlation (DCC), GDS, tensor canonical correlation analysis (TCCA), multilinear principal component analysis (MPCA), multilinear linear discriminant analysis (MLDA), PGM, and tangent bundle (TB). These methods are established in classification problems involving tensorial data and operate on subspace representations to classify tensor data. According to Table~\ref{tab:tab_related}, PGM and TB consistently produce competitive results. PGM produces reliable results compared to supervised methods, such as DCC and GDS. This observation is an indication of the advantages that the unfolding process employed by the tensor representation can provide.



\begin{table*}[htpb!]
\centering
\small
\caption{Cambridge and KTH datasets evaluation.}
\label{tab:tab_related}
\resizebox{0.95\linewidth}{!}{
\begin{tabular}{@{}lcc@{}}
\toprule
method                  & Cambridge~\citep{kim_dataset} & KTH~\citep{KTH} \\ \midrule
DCC~\citep{Kim_Tensor}   & 76\%  & 90\%  \\
GDS~\citep{GDS}          & 78\%  & 91\%  \\
TCCA~\citep{KIM_DCC2}    & 82\%  & 95\%  \\
MPCA~\citep{Tensor_PCA}  & 42\%  & 67\%  \\
MLDA~\citep{Tensor_LDA}  & 43\%  & 71\%  \\
PGM~\citep{LUI2}         & 88\%  & 96\%  \\
TB~\citep{TB}            & 91\%  & 96\%  \\
$n$-mode GDS             & 93\%  & 97\%  \\
$n$-mode wGDS            &\bf 94\%  &\bf 98\%  \\ \bottomrule
\end{tabular}}
\end{table*}

\subsection{Comparison with Existing Methods using Handcrafted Features}


In this experiment, we evaluate the proposed method with state-of-the-art handcrafted features and compare it with methods related to deep learning. For this comparison, we employ the 3D convolutional neural network (C3D)~\citep{3dcnn}, two-stream network~\citep{2stream}, and two-stream network I3D~\citep{2streamI3D}. The C3D is equipped with spatio-temporal three-dimensional kernels, improving the performance levels in the field of action recognition. In contrast, the two-stream network learns different types of features which are combined for action classification. In this network, a spatial-CNN is trained to extract appearance features using RGB images, while a temporal-CNN is trained using optical flow to extract the motion pattern. The stream features are then concatenated to represent actions in realistic videos.
Although deep learning approaches have made significant advances in video-related tasks, handcrafted features produce competitive results compared with the state-of-the-art on many standard action recognition tasks~\citep{feat_sota1,feat_sota2}. These solutions are usually based on improved variations of HOG and HOF, for instance, improved dense trajectories (iDT)~\citep{idt0}.

The proposed $n$-mode GDS can benefit from handcrafted features since the $n$-mode subspace representation allows the use of types of features other than raw features. In this section, we investigate the combination of $n$-mode GDS and handcrafted features, connecting the best of both strategies via a single hybrid tensor classification architecture. To evaluate the synergy between $n$-mode GDS and handcrafted features, we use HOG, HOF, MBH, and iDT. These handcrafted features have different characteristics that are exploited in the proposed method. HOG is able to extract the local appearance and shape of objects by using local intensity gradients. In this experiment, the HOG features replace the $3$-mode unfolding, since this mode comprises the appearance of the actions. HOF is a popular handcrafted feature that can accurately obtain the motion information from videos. Although it is similar to HOG, HOF includes optical flow data across the frames, preserving temporal information. In this experiment, the HOF descriptor replaces both $1$- and $2$-mode unfolding. Finally, the MBH descriptor works by extracting the derivatives of the horizontal and vertical components of the optical flow. MBH preserves the relative motion between pixels and represents the gradient of the optical flow. The camera motion is removed, and information about changes in the motion boundaries is preserved. As a result, MBH is robust to camera motion and provides discriminative features for action recognition.





\begin{table}[]
\centering
\small
\caption{The average accuracy of $n$-mode GDS and deep learning approaches.}
\label{tab:dnn-results}
\begin{tabular}{@{}lcc@{}}
\toprule
\multicolumn{1}{c}{\multirow{2}{*}{approaches}} & \multicolumn{2}{c}{datasets}           \\ \cmidrule(l){2-3} 
\multicolumn{1}{c}{}         &  HMDB-51 &  UCF-101          \\ \midrule
C3D~\citep{3dcnn}                         &  51.9    &  85.4             \\
Two-stream~\citep{2stream}                   &  59.4    &  88.3             \\
Two-stream I3D~\citep{2streamI3D}               &  \bf80.7 & \bf98.0           \\  
$n$-GDS                      &  45.1    &  73.6             \\
$n$-wGDS                     &  46.6    &  75.7             \\
$n$-wGDS + HOG               &  48.3    &  77.1             \\
$n$-wGDS + HOF               &  51.8    &  80.2             \\
$n$-wGDS + MBH               &  53.5    &  82.7             \\
$n$-wGDS + iDT               &  55.7    &  83.9             \\ \bottomrule
\end{tabular}
\end{table}

In the last experiment, we use the improved trajectory features (iDT), which is a state-of-the-art handcrafted descriptor proposed by~\citep{idt0} for human action recognition. This robust descriptor employs HOG, HOF, and MBH, followed by dimensionality reduction and Fisher vector encoding. Since the above descriptors are based on a 1-dimensional histogram representation of individual features (HOG, HOF, MBH), they directly model values of given features and can be directly employed as subspaces in the proposed framework. Table~\ref{tab:dnn-results} lists the results of I3D, C3D, and two-stream convolutional networks, as well as the results attained by the proposed method and its combination with handcrafted features commonly used in the literature. According to the results, the $n$-mode GDS improves the results when equipped with the weighted geodesic distance while extracting features from the modes improves its accuracy even further. The use of HOG features as the appearance mode improves the $n$-mode GDS accuracy by about $2\%$ in both datasets, confirming that extracting appearance features is beneficial for the proposed method.

While I3D provides the best results in these experiments, the training required for this deep neural network is preventive for more specific applications. For instance, I3D is pre-trained on ImageNet and Kinetics Human Action dataset. However, the proposed method does not rely on pre-training, making use of handcrafted features, which covers a broader range of applications. The use of HOF and MBH increases the $n$-mode GDS accuracy by about $7\%$. These features provide temporal information, which is an advantage comparing to HOG features. The discriminative approach combined with the weighting strategy employed by the proposed method gave competitive results for the $n$-mode GDS. By employing all the available descriptors, $n$-mode GDS projection produces competitive results compared with C3D and two-stream convolutional networks, confirming the effectiveness of the proposed method.

In this experiment, we can see the effects of $w$ over $n$-mode GDS performance. When the weight approach based on the $n$-mode Fisher score is employed, the proposed method is improved by $1.5\%$ and $2\%$ in both datasets. The weights computed by the $n$-mode Fisher score generated $0.33$, $0.30$, and $ 0.35$ for the 1-, 2-, and 3-mode subspaces, respectively. The approach appears to be very efficient mostly because it enables a direct estimation of the importance of each mode directly, without the use of an exhaustive search. When HOG features are employed to replace the raw images of the 3-mode subspace features, the $n$-mode Fisher score estimates the weights as $0.32$, $0.29$, and $0.38$ for the 1-, 2-, and 3-mode subspaces, respectively. This new set of weights increases the importance of the tensor mode, which uses the handcrafted features, implying the importance of these features for the tensorial class separability and consequently to the classification accuracy. This behavior is found in other experiments, where the tensor mode employing handcrafted features is reported with higher weights than other modes. These observations confirm the hypothesis that interpreting spatial-temporal modes across the manifolds is useful, and how to preserve a balance between these modes is essential for improving the accuracy of the proposed method.
















\section{Conclusion and Future Directions}
\label{S:Conclusion}

In this paper, we proposed a tensor representation and classification method based on PGM. By exploiting the geometry of the product manifold through its metric, the proposed $n$-mode GDS projection based on the subspace learning method obtained a discriminative model on the PGM. The $n$-mode Fisher score was also proposed to evaluate the $n$-mode subspace separability of the new model. We also introduced the weighted geodesic distance into the proposed model. 

The proposed method was investigated in action and gesture recognition problems. The high performance in the classification experiments conducted on different video datasets indicates that the new model is well suited to representing high dimensional data and revealing intrinsic subspaces structures underlying the data. In future work, we intend to focus on investigating different metrics of the PGM and testing the proposed methods on larger scale complex videos. Another possible research direction is to extend this framework to take into account the nonlinear nature of the data distribution, such as by employing a kernel approach to handle nonlinear patterns~\citep{mkernel,mkernel2}.

\section*{Availability of Data and Material}

KTH~\citep{KTH}, Cambridge Gesture~\citep{kim_dataset}, HMDB-51~\citep{action51}, UCF-101~\citep{action101} and The Osaka University Kinect Action Dataset~\citep{Osaka} are used for training and testing the proposed method. These datasets are freely available. The source code of the proposed method is available at: \url{https://github.com/bernardo-gatto/n-mode-GDS}.

\section*{Competing Interests}
The authors declare that they have no competing interests.

\section*{Credit Authorship Contribution Statement}
\textbf{Bernardo Bentes Gatto}: Conceptualization, Methodology, Software, Validation, Formal Analysis, Investigation, Data Curation, Writing - Original Draft, Visualization, Funding Acquisition. \textbf{Eulanda Miranda dos Santos, Alessandro Lameiras Koerich, Kazuhiro Fukui and Waldir Sabino da Silva J\'{u}nior}: Writing - Review \& Editing, Supervision.

\section*{Acknowledgment}
This work was supported by JSPS KAKENHI Grant Number 19K20335.









\bibliography{main}

\end{document}